\documentclass[sigconf]{acmart}

\usepackage{booktabs} 

\usepackage{booktabs}
\usepackage{multirow}
\usepackage{bbm}
\usepackage{enumitem}
\usepackage[justification=centering]{caption}
\usepackage[flushleft]{threeparttable}

\copyrightyear{2018} 
\acmYear{2018} 
\setcopyright{acmcopyright}
\acmConference[WSDM 2018]{WSDM 2018: The Eleventh ACM International Conference on Web Search and Data Mining }{February 5--9, 2018}{Marina Del Rey, CA, USA}
\acmBooktitle{WSDM 2018: WSDM 2018: The Eleventh ACM International Conference on Web Search and Data Mining , February 5--9, 2018, Marina Del Rey, CA, USA}
\acmPrice{15.00}
\acmDOI{10.1145/3159652.3159683}
\acmISBN{978-1-4503-5581-0/18/02}

\begin{document}

\title[Modeling Time to Open of Emails]{Modeling Time to Open of Emails with a Latent State for User Engagement Level}

\author{Moumita Sinha}
\affiliation{%
    \institution{Adobe Research}
}
\email{mousinha@adobe.com}

\author{Vishwa Vinay}
\affiliation{%
    \institution{Adobe Research}
}
\email{vinay@adobe.com}

\author{Harvineet Singh}
\affiliation{%
    \institution{Adobe Research}
}
\email{harvines@adobe.com}

\begin{abstract}

Email messages have been an important mode of communication, not only for work, but also for social interactions and marketing. When messages have time sensitive information, it becomes relevant for the sender to know what is the expected time within which the email will be read by the recipient. In this paper we use a survival analysis framework to predict the time to open an email once it has been received. We use the Cox Proportional Hazards (CoxPH) model that offers a way to combine various features that might affect the event of opening an email. As an extension, we also apply a mixture model (MM) approach to CoxPH that distinguishes between recipients, based on a latent state of how prone to opening the messages each individual is. We compare our approach with standard classification and regression models. While the classification model provides predictions on the likelihood of an email being opened, the regression model provides prediction of the real-valued time to open. The use of survival analysis based methods allows us to jointly model both the open event as well as the time-to-open. We experimented on a large real-world dataset of marketing emails sent in a 3-month time duration. The mixture model achieves the best accuracy on our data where a high proportion of email messages go unopened. 

\end{abstract}

\keywords{Email interaction data, survival analysis, time-to-event prediction, enterprise email marketing, Cox-proportional hazards model}

\maketitle

\section{Introduction}

Email has a rich history of being a data source for machine learning techniques. Starting with spam filtering \cite{emailSpam}, the range of applications today covers a rich spectrum of scenarios. 
The Enron Corpus \cite{klimt2004enron} enabled research into the modeling of users' interactions with email in a collaborative environment \cite{chapanond2005graph}. For email service providers, detailed understanding of consumers' interactions with the email system allows building predictive models for specific actions, e.g. if an email will be replied to or not \cite{yang2017} and creating rich experiences \cite{advancedEmailFeatures, smartreply} for the recipients. On the consumer side, given its popularity, there has been much work on different ways to handle large volumes of email effectively \cite{whittaker1996email}. An early paper by Horvitz et al. \cite{horvitz1999attention} proposed that autonomous agents may be able to identify and prioritize emails that need attention. The authors of \cite{predictEmailActions} show that historical data allows the prediction of what actions a user might take on the receipt of an email, for example, marking it for deletion \cite{dabbish2003marked}. Apart from being a mode of communication, email is also used as a personal information management environment \cite{ducheneaut2001mail}, leading to the need to support other forms of interactions like search \cite{emailSearch}.

The domain of interest in the current paper is marketing, where the email channel ranks high in popularity \cite{vanboskirk2011us} alongside social media, search \& display advertising. 
Email based marketing is predicted to have a compound annual growth rate of $10\%$ \cite{vanboskirk2011us} and nearly every enterprise marketer uses it as a delivery channel \cite{britishairways, mint}. The engagement levels however are typically low, as compared to personal email messages at work or among friends. The open rates for the marketing email messages, vary by industry - ranging from $15\%$ to $19\%$ in the e-commerce, beauty and personal care, and gambling industries, and in the range of $20\%$ to $28\%$ in the hobbies, home and garden or health and fitness industries \cite{openrate}. Marketers are therefore always on the lookout for techniques that might enhance the engagement levels. For example, Kumar et al. \cite{kumar2014modeling} modeled opt-in and opt-out behaviour and related these to transactions made by the consumer. Bonfrer et al. \cite{bonfrer2009real} proposed a framework that allows real-time evaluation of an email campaign.

In this submission, we propose the use of \textit{survival analysis} for jointly modeling the open event on an email, as well as the time-to-open. The next section provides technical background to some important concepts in survival analysis that are relevant in the current scenario.

\section{Survival Analysis}
\label{secSurvivalAnalysis}

\begin{figure}[t]
    \centering
    \includegraphics[width=0.9\columnwidth]{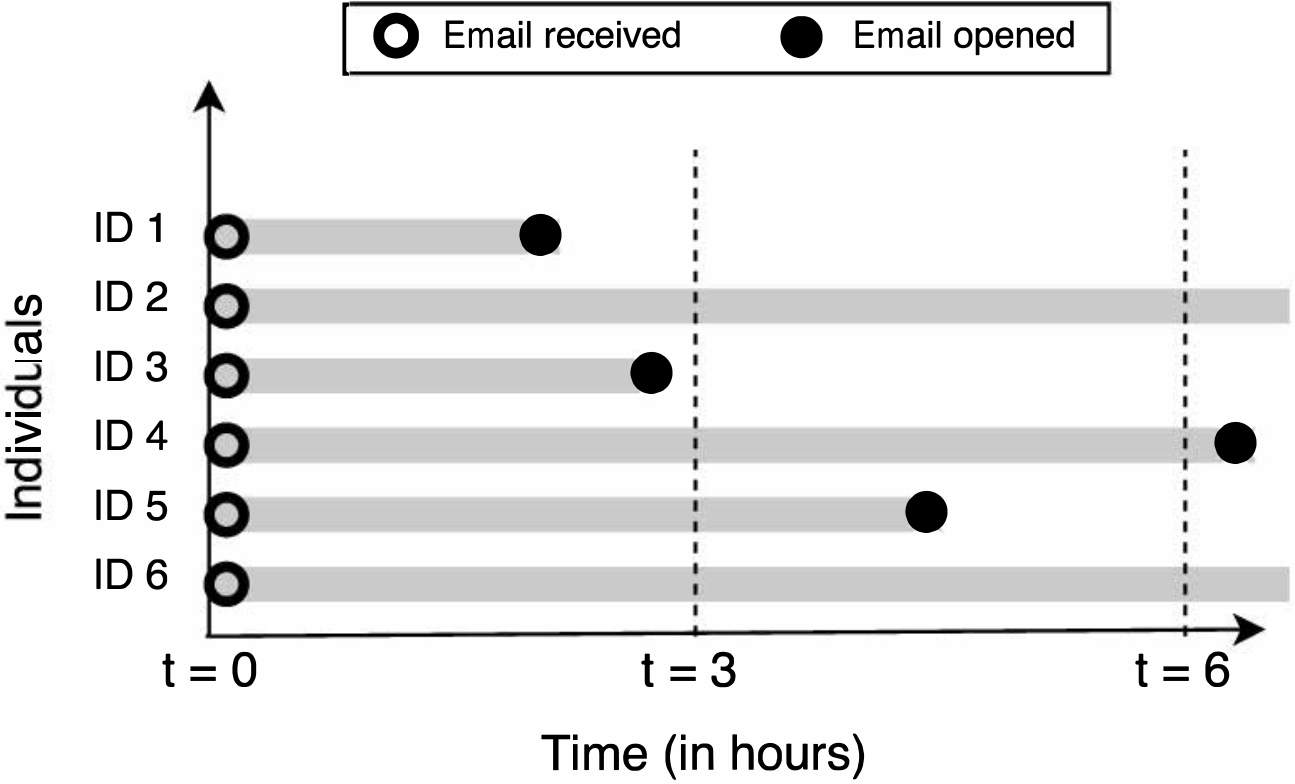}
    \caption{Example of survival data with different censoring windows. Emails are received at t=$0$ by all individuals. Note that at t=$3$ hours, outcome for individual ID $5$ is censored but will be observed for t=$6$ hours.}
    ~\label{fig:survival_data}
\end{figure}

Survival analysis refers to an area of statistical modeling where the main variable of interest is the time to an \textit{event}. Historically, the event is assumed to be \textit{death}.
One characteristic of data that makes the use of survival models appropriate is the presence of \textit{censoring}. This refers to the fact that not all individuals would have experienced the event within the observation window. The censoring may be because at the time of analysis the event had not yet occurred, or if the corresponding individual can no longer be tracked. Figure~\ref{fig:survival_data} is a pictorial representation of survival data in the context of emails. Observations are synchronized at $t=0$, which is the time at which the individuals receive the email. If the event of the email being read is not within a chosen time interval, e.g. $t=3$ hours, this would be a censored data point. And some recipients may of course not read the email at all. 

Consider a random variable $T$ for the time to the event of interest, with the corresponding probability density function $f(t)$ and the cumulative distribution function being $F(t)$ at a given time $t$. Then the survivor function is defined as
\begin{equation}
S(t) = P(T\geq t)=1-F(t)=\int\limits_{t}^{\infty} f(u) du
\end{equation}

It represents the probability that an individual will survive beyond time $t$. Equivalently, given that the individual has not yet experienced the event till time $t$, the hazard function $h(t)$ represents the instantaneous chance of the event occurring at time $t + dt$.

\begin{equation}
\label{cox-ph}
    h(t)=\lim_{dt\to 0} \frac{P(t \leq T <t+ dt| T \geq t)}{dt}
\end{equation}

The relationship between the survivor function and the hazard function can be derived as being $S(t) = exp\big\{-H(t)\big\}$, where $H(t)$ is the cumulative hazard function corresponding to $h(t)$.

A survival analysis dataset containing \textit{N} individuals is represented as $\{X_i, Y_i, \delta_i\}$, with $i=1 \dots N$.  For the $i^{th}$ individual, $X_i$ is a vector of features that are believed to be predictive of the survival time. The target $Y_i=min(t_i, C_i)$ represents the survival time, where $C_i$ represents the duration of time for which the individual was observed and is also known as the censoring window. If observed within the censoring window, $t_i$ is the time to event for the $i^{th}$ individual. The indicator variable $\delta_i$ encodes if the $i^{th}$ individual experienced the event of interest within the censoring window. 

\begin{equation}
  \label{eqdelta}
  \delta_i=\begin{cases}
    1, & \text{if $t_i < C_i$}\\
    0, & \text{otherwise}
  \end{cases}
\end{equation}
  
\subsection{Cox Proportional Hazard Regression}
Given a feature vector $X_i$ for the $i^{th}$ individual, the hazard function for the individual at any given time $t$ can be defined as 
\begin{equation}
\label{cox-ph-proportion}
h_i(t|X_i)=h_0(t) \times \psi(X_i)
\end{equation}

Here $h_0(t)$ is the baseline hazard function at time $t$, and $\psi(.)$ incorporates the dependence on the individual-specific features $X_i$, which are independent of time. The specific factorization of $h_i(t|X_i)$ into a global time-dependent component ($h_0(t)$) and an individual's time-independent factor ($\psi(X_i)$) is the \textit{Proportional Hazards} assumption - Section \ref{secModelAsssumptions} provides a methodology to validate this assumption on a given dataset. What has been defined above is a semi-parametric approach, in that no assumptions have been made about the shape of the baseline hazard function $h_0(t)$. The parametric alternative would be to impose a functional form, e.g. a Weibull distribution. Based on the relation between the survivor and hazard functions, the survivor function of the $i^{th}$ individual for Cox Proportional Hazard (CoxPH) regression is 
\begin{equation}
\begin{split}
  S_i(t|X_i)& = exp\big\{-\int\limits_{0}^{t} h_0(u)\psi(X_i) du\big\}\\
            & =[exp\big\{-H_0(t)\big\}]^{\psi(X_i)}\\
            & =S_0(t)^{\psi(X_i)}
\end{split}
\end{equation}

The corresponding partial likelihood function \cite{cox1972regression} is defined as
\begin{equation}
\label{likelihood}
 L(\beta)= \prod \limits_{i : \delta_i = 1} \frac{\psi(X_i; \beta)}{\sum \limits_{l \in R(t_i)}\psi(X_l; \beta) }  
\end{equation}

where the function $\psi(.)$ has been parameterized by $\beta$ that controls the combination of the features. $R(t_i)$ is the set of individuals who are \textit{at-risk} of the event at time $t_i$, that is, the set of individuals for whom the event has not occurred yet. $t_i$ is also the observed time to event of the $i^{th}$ individual. Note that the numerator of the likelihood is a function of only the individuals that observed the event, and censored individuals only contribute to the denominator of Equation \ref{likelihood}. The $\beta$ values are estimated by maximizing the above likelihood using a gradient based method.

The most common form of $\psi(X_i) = exp\big(\beta^TX_i\big)$, where $\beta$ is a vector of parameters controlling the dependence between the features in $X_i$ and target $Y_i$. Doing so assumes a linear scaling of the relative (log) hazards of different individuals with respect to the values of the features. Ridgeway \cite{gbmCPH} proposed that the likelihood in Equation ~\ref{likelihood} can alternatively be optimized directly using gradient boosting methods that might provide benefits in scenarios where the effect of the features is non-linear. Note that this is still a Proportional Hazards model, but with $\psi(X_i)$ taken to be the output of a gradient boosting machine (GBM).

\subsection{Mixture Model with Cox Proportional Hazard Regression}
The CoxPH model assumes that all individuals will eventually experience the event. But there may be a proportion of individuals who are not prone to the event, i.e., who are not predisposed to opening emails. The level at which an individual user is engaged with marketing messages influences his/her act of opening the email (and how quickly). The CoxPH model described earlier tries to explain all the observations using only the features ($X_i$) as the explanatory factors. Through the use of mixture models \cite{farewell1982use, branders2015mixture}, we might expect to get more discriminatory power. The $i^{th}$ individual is now represented as $\{X_i,Y_i, \delta_i, L_i, Z_i\}$, where $L_i$ is a latent indicator variable such that

\begin{equation}
  L_i=\begin{cases}
    1, & \text{prone to the event}\\
    0, & \text{otherwise}
  \end{cases}
\end{equation}

$Z_i$ is a set of features that help predict if an individual is prone to the event of interest or not. The feature set $Z_i$ can also be the same as the feature set $X_i$.

\begin{equation}
  \pi(Z_i)=P(L_i=1|Z_i)=\frac{exp\big(\mathbf{b}^TZ_i\big)}{1 + exp\big(\mathbf{b}^TZ_i\big)}
\end{equation}

The probability $P(L_i=1|Z_i)$ is estimated using logistic regression here, and is introduced as a mixture probability into the overall survivor function:
\begin{equation}
\label{eqMM}
  S_i(t|X_i)=\pi(Z_i)S(t_i|L=1,X_i) + (1 - \pi(Z_i))
\end{equation}
If the individual is predisposed to not experiencing the event, then $\pi(Z_i) \simeq 0$, leading to a prediction of a survival probability close to $1$. Conversely, a scenario with $\pi(Z_i) \simeq 1$ leads to the first term dominating, with the quantity $S(t_i|L=1,X_i)$ representing the survival probability in the traditional sense. A proportional hazards assumption can be encoded by setting $S(t_i|L=1,X_i) = S_0(t)^{exp\big(\beta^TX_i\big)}$ as before. The likelihood of the model is given by:

\begin{multline}
  L(\beta,b)=\prod\limits_{i= 1}^{N} \left( [1-\pi_i(Z_i)]^{1-L_i} \right. \times\\
  \left. [\pi_i(Z_i)S(t_i|L_i=1,X_i)\{h(t_i|L_i,X_i)\}^{\delta_i}]^{L_i} \right)
\end{multline}

Since there are latent variables (the $L_i$), the optimization is an Expectation Maximization based iterative procedure that estimates the $L_i$, along with $\mathbf{b}$ (for calculating $\pi(Z_i)$) and $\beta$ controlling how the features of an individual affect the relative hazards. In the current setting, we are interpreting $\pi_i(Z_i)$ as the engagement level of a given user $i$, the model however is more general. For e.g., it can be used to represent the probability that a patient has been cured, which in turn affects the chances that he/she will experience the event.

\subsection{Related Work}
\label{secRelated}

Survival analysis has traditionally been used in the health-care domain to determine the time to `death' in patients, but the usage of this range of techniques has recently expanded to other application areas \cite{wangmachine}. Examples include prediction of early student dropouts \cite{ameri2016survival}, post-click engagement on native ads \cite{barbieri2016improving}, query specific micro-blog ranking for improved retrieval \cite{efron2012query}, recommender systems in e-commerce \cite{wang2013opportunity}, search engine evaluation via the use of "absence time" \cite{chakraborty2014}, and predicting time for crowd-sourced tasks \cite{lease2011crowdsourcing}. 

By appropriately defining the event being modeled, existing marketing concepts also lend themselves survival analysis techniques. E.g. re-purchasing behavior is an indicator of high engagement \cite{leesurvival} and a proxy for the potential value of a customer \cite{drye2001customers, lu2003modeling}.
Attrition modeling helps businesses identify customers who are most at-risk so that attempts can be made to keep them in the system, and \cite{leesurvival} proposes a survival analysis based solution. 

Much of the literature referred to above involve applying well-known and established models (like CoxPH) in different scenarios. But more recently, growing interest in the use of survival analysis has led to modeling improvements. For instance, when modeling time-to-event of related tasks, the parameters of the different models can be more reliably estimated using regularization techniques commonly used in multi-task learning \cite{li2016multi}. Even in traditional application areas of survival analysis, given a large number of data points and a variety of features that potentially have a highly non-linear dependence on the time-to-event, deep latent models provide better performance \cite{ranganath2016deep}.

The closest related work to that presented here is described in \cite{dave2017fast} where time-to-event is modelled in the email domain. Given this context, the contribution of the current paper is two-fold: (1) we describe techniques from the rich history of survival analysis to identify those models whose assumptions are better matched with the characteristics of the data (2) for the application of predicting time-to-event when the censored rows dominate, the mixture model (MM) described above is shown to not only describe the data better but also provide better predictive performance.

\section{Problem Definition and Data Description}

When emails containing time sensitive information are sent, it may be relevant for the sender to know what is the expected time within which the email will be read by the recipient. Specifically in marketing messages, if the email advertises a flash sale, the marketer will need to decide on the time window for the sale - to optimize between reaching sufficient consumers within the window and yet keep it exclusive. Prediction of time-to-open of an email by a consumer helps to determine the size of the recipient list one wants to reach.

Our dataset corresponds to email marketing campaigns that are sent out to consumers of an enterprise and 
we are interested in a predictive model that answers questions of two types: (a) Is a particular email likely to be opened by a given recipient? (b) Can we predict the time within which the email will be opened?

In the dataset, there is a high degree of variability amongst the marketing messages - some are sent to a large group of recipients, while others are targeted at a narrow set of consumers -  e.g. a personalized birthday communication. We expect that the nature of people's interaction with these different types of emails varies drastically. In particular, we are interested in modeling how people differ in terms of their engagement with the mass marketing emails. For this reason, the analysis presented here includes only those emails that were sent to at least $50\%$ of the total consumers. We have additionally dropped those consumers who received fewer than $10$ messages during the period of interest.

The time at which an email reaches a consumer is labelled as its \textit{start-time}. In the event that the email is read, the email has a corresponding \textit{open-time}. The difference between the two time-stamps is referred to as the \textit{time-to-open}. The emails are divided into 3 non-overlapping buckets based on the start-time: a \textbf{Training} dataset (spanning 4 weeks) and one dataset each for \textbf{Validatation} \& \textbf{Test} (spanning 3 weeks each) respectively. Table \ref{tbldataStats} shows the size of each of these datasets. Chronologically ordered, these 3 datasets cover 13 weeks of email messages with a $3$ week gap between \textbf{Validatation} and \textbf{Test}. Within each group, data from the initial two weeks are used to compute features that will be used to model users' interaction with emails sent in the subsequent week(s).

\begin{table}[!htbp]
\centering
\caption{Overview of Datasets}
\label{tbldataStats}
\begin{tabular}{ |c|c|c| } 
 \hline
 \textbf{Dataset} 
 & \multicolumn{1}{|p{2cm}|}{\centering \textbf{\#Recipients} \\ (millions)} 
 & \multicolumn{1}{|p{2cm}|}{\centering \textbf{\#Emails} \\ (millions)} \\
 \hline
 \textbf{Training} & 2.05 & 31.86 \\
 \textbf{Validation} & 2.04 & 22.73 \\
 \textbf{Test} & 2.22 & 19.14 \\
 \hline
\end{tabular}
\end{table}

\subsection{Survival and Censoring}
\begin{figure}[t]
    \centering
    \includegraphics[width=0.75\columnwidth,height=0.75\columnwidth,keepaspectratio]{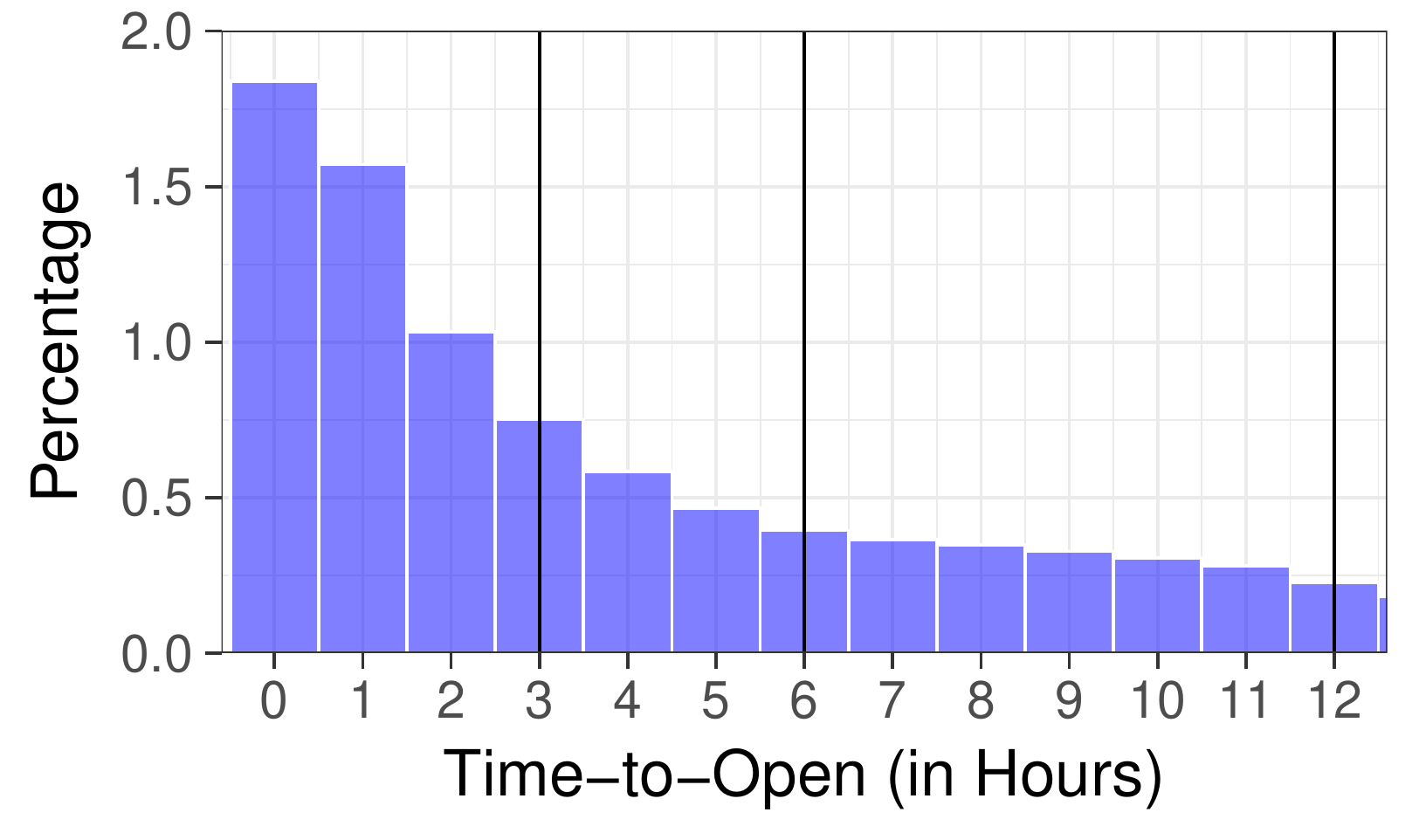}
    \caption{Distribution of time-to-open as a percentage of the number of recipients}
    ~\label{fig:time_to_open}
\end{figure}

In the context of email messaging, the opening of an email is the event and the survivor function at time $t$ provides the probability that the email will not be opened by this time. For the purposes of the results presented here, the data was gathered by monitoring the emails for $10$ days after they were sent. Figure \ref{fig:time_to_open} shows the distribution of the time-to-open in our \textbf{Training} dataset. Note that the plot accounts for only $\sim8\%$ of the total data points.

In applications of survival models, the size of the censoring window is dictated by when the analysis needs to be performed or when information stops being available about the individuals. The choice of censoring window should include as many of the eventual opens as possible. Amongst the emails that were opened within the observed 10 days, $\sim43\%$ happened within the first $3$ hours, with the corresponding numbers for $6$ and $12$ hours being $\sim57\%$ and $\sim74\%$ respectively. Beyond $12$ hours, the opens become very rare. 

With $C_i=3$ hours for each individual $i$, over half the eventual opens would be lost - this is undesirable. On the other hand, there are practical limitations to being able to track and monitor an email campaign for extended periods of time. In addition, for time sensitive messages, the censoring window is the time within which the message needs to be opened by the recipients to ensure that the promotion is valid. Given the observations above about the data and the constraints imposed by the domain, in the current paper, we are evaluating three different censoring windows - $3$, $6$ and $12$ hours - with a preference for the higher values. For a given choice of $C_i$, the emails that were not opened by their recipients within the time period are considered censored.

Given that all individuals are synchronized at $t=0$ (Figure \ref{fig:survival_data}), the $C_i$'s across all individuals will be the same for a given choice of censoring window. Note that when the censoring window changes, the training data provided to the survival models changes (via Equation \ref{eqdelta}). Therefore, identifying the \textit{right} censoring window for the given application requires us to build a given model multiple times - one for each value of $C_i$.

Across the dataset, we can construct a histogram of the cumulative fraction of individuals that \textit{survived} (i.e., emails that are yet to be opened) up until a given time. The Kaplan-Meier estimator \cite{kaplan1958nonparametric} constructs a smooth non-parametric approximation to this histogram. In Figure ~\ref{fig:mm_survplot}, the curve corresponding to ``Combined'' represents the \textbf{Training} dataset, the significance of the other data in the plot will be explained in Section ~\ref{secModelAsssumptions}. We can see how the curve has a survival rate of $95\%$ at around $3$ hours - this represents the fact that only $5\%$ of the users have actual time-to-open data for the censoring window of three hours (as also seen in Figure ~\ref{fig:time_to_open}). This high degree of censorship is a unique characteristic of our dataset and represents one of the challenges in terms of the traditional use of survival models.

\subsubsection{Baselines: }
We consider the historical open rate of an individual as the baseline for predicting his/her future open events. We also include a Logistic Regression (LR) model as a classification baseline in our experiments. The auxiliary problem of predicting the time to the open event is not as straightforward. As a simplistic baseline, we consider the censoring time $C$ as the predicted time to open for all individuals (B). Given that the majority of rows are censored, $C$ is taken as the median of the $Y_i$ values, with the mean being only marginally lower. We also considered a linear regression (LR) model to predict time to open, as a baseline in our experiments which excluded all individuals who had censored information from the model \cite{burke1997artificial, delen2005predicting}. Here, with slight abuse of notation, we denote both logistic and linear regression baselines by LR. The distinction is made clearly while presenting results.

\subsection{Testing Model Assumptions}
\label{secModelAsssumptions}
In this section, we will describe statistical analyses that validates the choice of  models in our experimental section. Every model represents a corresponding set of statistical assumptions, and we need to verify that the data aligns with the assumptions of the models used.

In the original CoxPH model, the hazard of an individual at a time $t$ is a product of two quantities - the baseline hazard and the individual's relative hazard (See Equation \ref{cox-ph-proportion}). The baseline hazard is a global quantity that incorporates the dependence on time. And the relative hazard of an individual is expressed as a parameterized function of features and is independent of time. Verifying the applicability of a proportional hazards model equates to validating that the features are independent of time. To test this independence, we take all observed (i.e., not censored) events for individual $i$ and calculate the scaled Schoenfeld residual \cite{grambsch1994proportional} for each feature $j$ as follows:
\begin{equation}
    r_{ij}= X_{ij}-a_{ij}
\end{equation}
where $X_{ij}$ is the value of the feature and 
\begin{equation}
a_{ij}=\frac{\sum_{k \in R\small(t_i\small)}X_{kj}exp\big(\beta^TX_k\big)}{\sum_{k \in R\small(t_i\small)}exp\big(\beta^TX_k\big)}
\label{eqSchoenfeld}
\end{equation}

\begin{figure}[t]
    \includegraphics[width=\columnwidth]{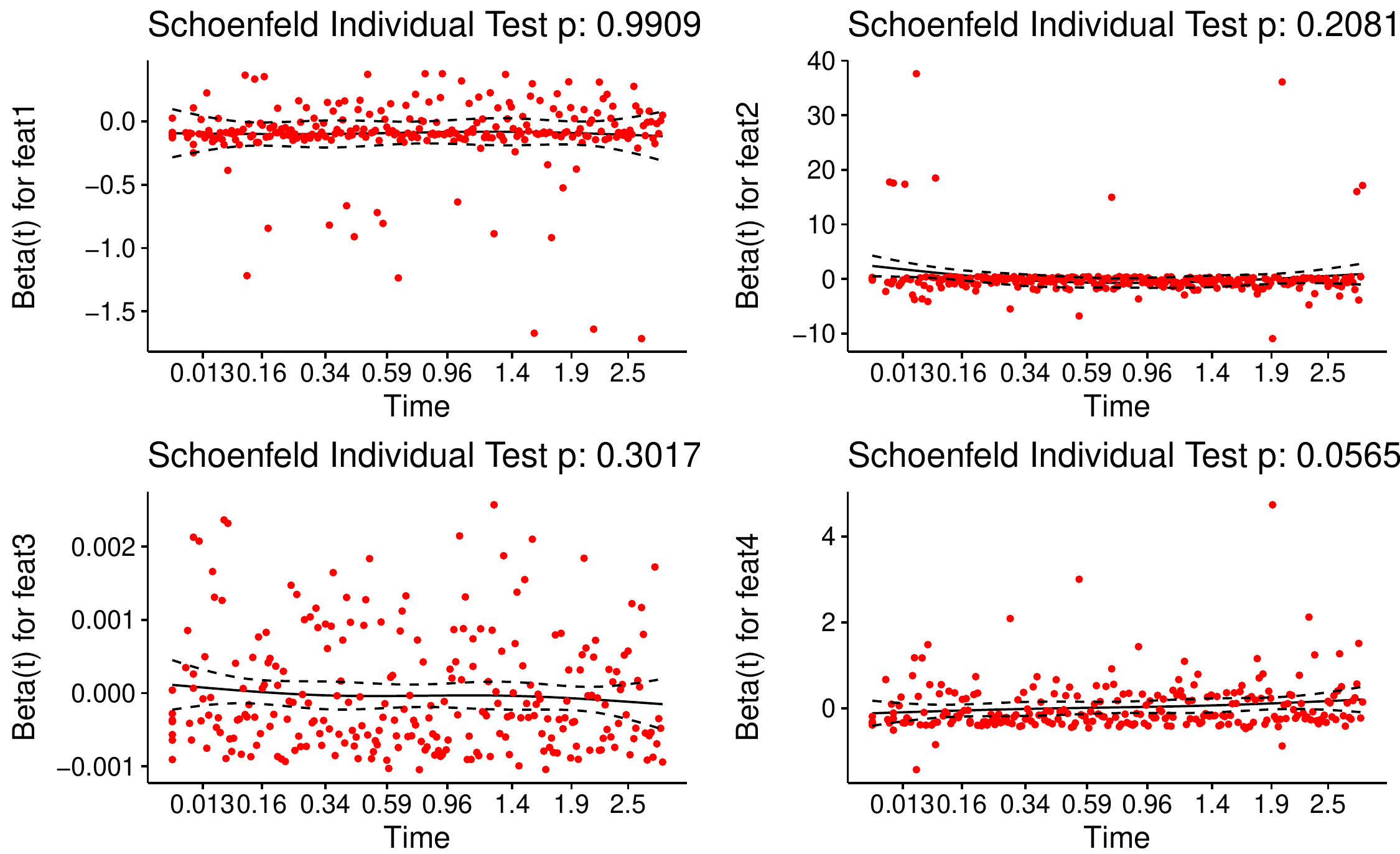}
    \caption{Test for the Proportionality Assumption in CoxPH for different features. The scaled Schoenfeld residuals do not have a trend over time}
    ~\label{fig:cox_assumption}
\end{figure}

In the equation above, $t_i$ represents the time at which the event occurred for individual $i$. And as before, the \textit{risk set} $R\small(t_i\small)$ represents those individuals who have not yet experienced the event until $t_i$. Plotting the values $r_{ij}$ as a function of the times $t_i$ allows us to validate the proportionality assumption - the check being that the $r_{ij}$ have no trend over time. Figure \ref{fig:cox_assumption} plots the scaled Schoenfeld residuals calculated using the cox.zph function in the \textit{survival} package in R. For the sake of clarity, only a sample of data points are displayed in each plot. Also included in the plot are the corresponding p-values for the null hypothesis test, that the sum of residuals for the corresponding feature across individuals is $0$. This procedure allows us to ensure that the features included in the model satisfy the null hypothesis (i.e., the proportionality assumption) at a chosen significance level - the features illustrated above have p-values greater than $0.05$.

Note that the above procedure utilizes the $\beta$ parameters from a previously fitted CoxPH model. A similar procedure can be used for the GBM (where the dependence between the features and the hazard is non-linear) as well as the Mixture Model (\textit{MM}). All the models considered in the current paper are based on the idea of \textit{proportional hazards}, hence the relevance of such a test.

We note two popular extensions available from the survival analysis literature that were excluded from our experiments because they did not match the characteristics of our scenario:
\begin{enumerate}[leftmargin=*]
    \item Parametric distributions, like the Weibull, are common choices for modeling event data. However, these were not appropriate here due to the multi-modal nature of the time-to-open, that is a large portion of the recipients are not prone to opening the message
    \item There exist extensions to the CoxPH model that are capable of utilizing time-dependent features. Given the nature of email data, where most messages that will be opened are opened within a relatively small period, obtaining updated predictions at short intervals due to changes in time-dependent feature values is not practical
\end{enumerate}

\begin{figure}[t]
    \includegraphics[width=0.8\columnwidth]{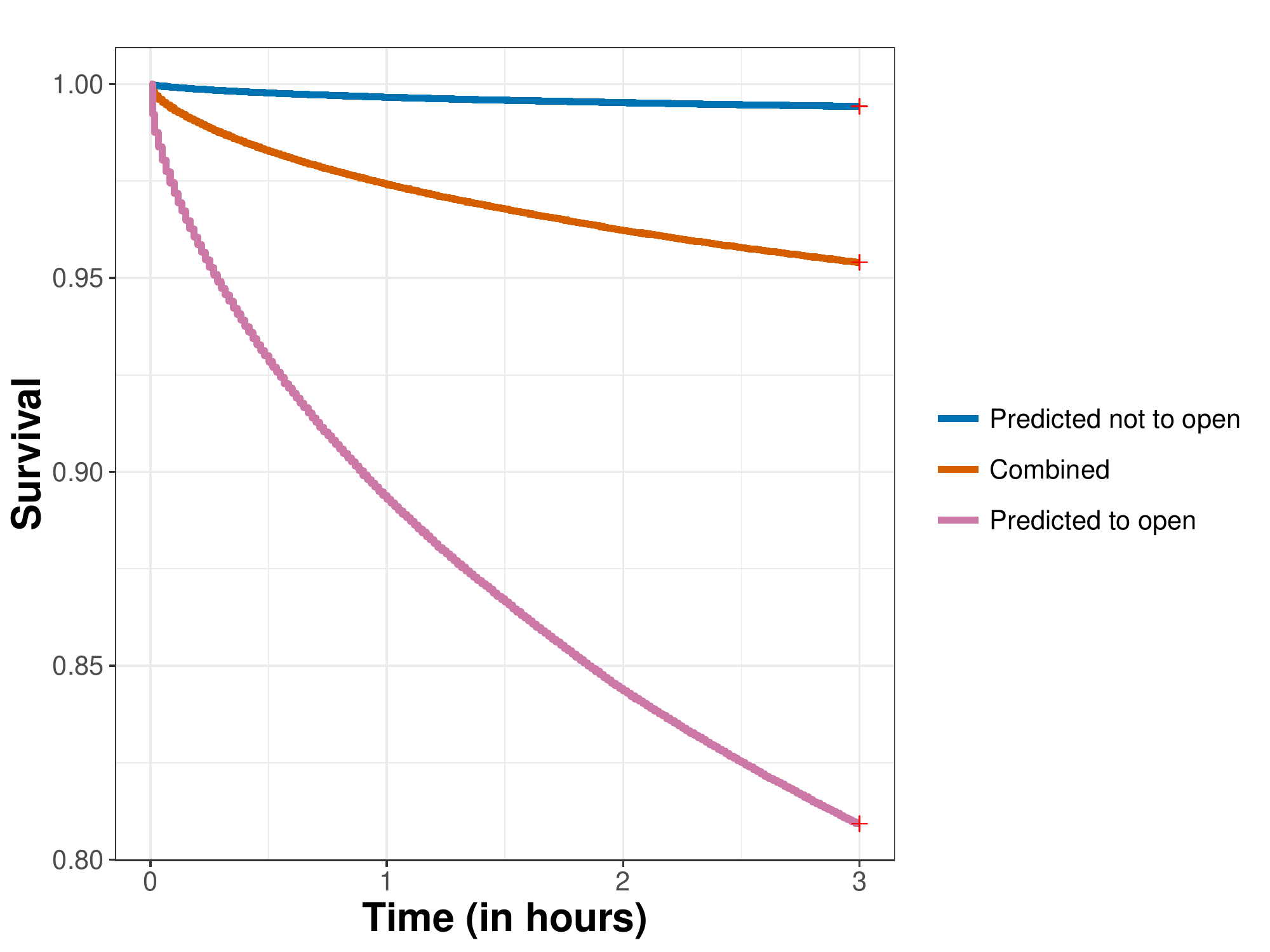}
    \caption{Survivor Functions of the Two Groups of Consumers}
    ~\label{fig:mm_survplot}
\end{figure}

Mixture models (MM) are applicable to scenarios where there are sub-populations with different characteristics. The MM estimates the probability of each individual belonging to one of the two latent states, those who will not open and those who may open the messages. The quantity $\pi(Z_i)$ in Equation ~\ref{eqMM}, therefore, represents how engaged the individual is with the email messages. A proxy method to test the existence of groups with differing engagement behaviors is to use a logistic regression model to classify the recipients into two groups - prone to opening their emails versus not - and plotting their survivor functions, as done in Figure \ref{fig:mm_survplot}. Statistical differences between the two groups can then be validated by a log-rank test, where the ranks of the survival times in the two groups are compared.

\subsection{Procedure for Predicting Time-to-Event}

The application scenarios discussed in the \textit{Related Work} section as well as the work described here, all utilize the main strengths of survival analysis - modeling time-to-event data as a function of features in the presence of censoring.

\begin{figure}[t]
    \includegraphics[width=0.8\columnwidth]{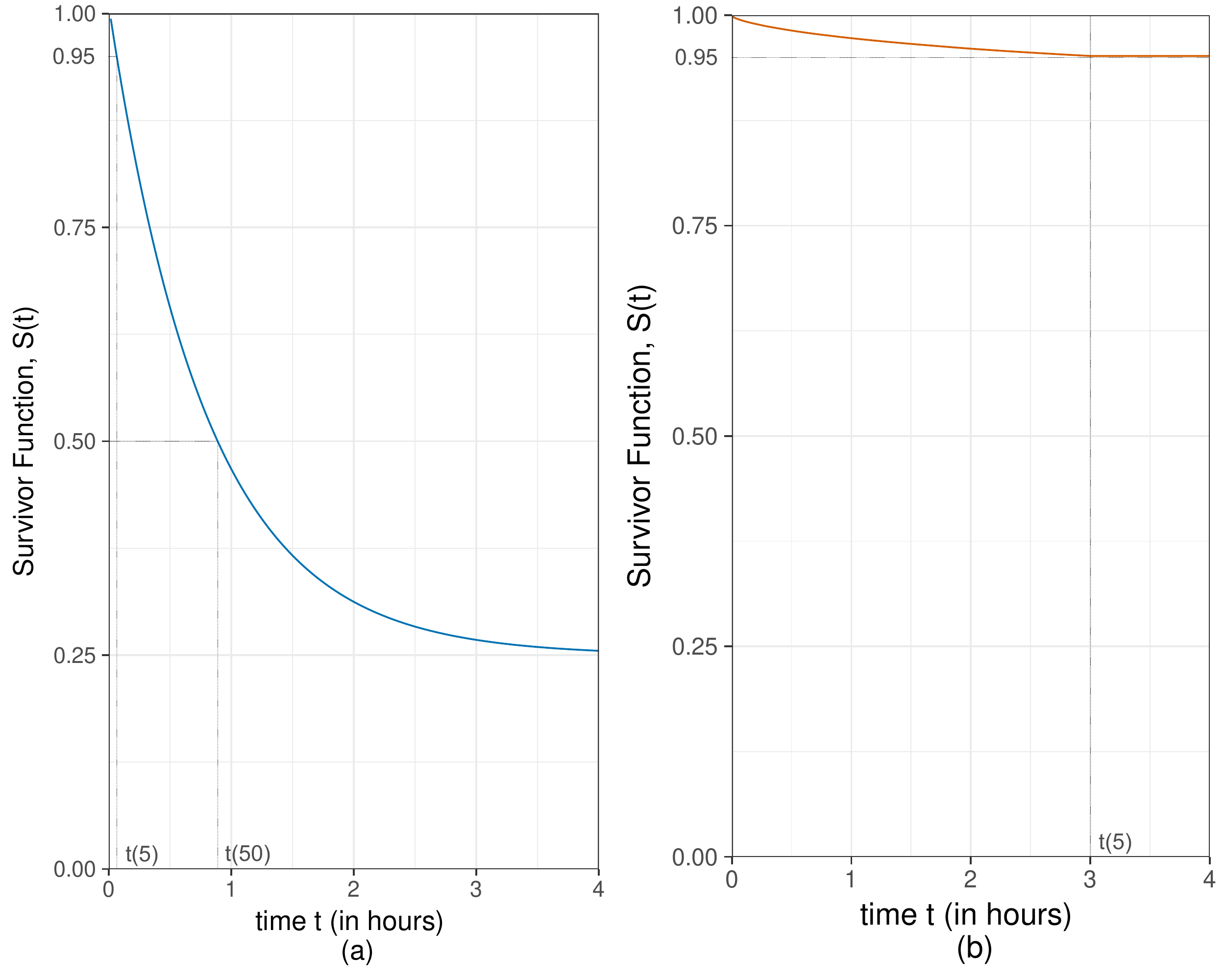}
    \caption{Two examples of survivor functions. (b) corresponds to our training data. Note that median survival time, i.e., t(50) is not defined in (b)}
    ~\label{fig:multiple_survplot}
\end{figure}

To predict the time-to-event for an individual (say $i$), we first calculate the hazard function for that individual. This is obtained by multiplying the estimate of the global baseline hazard, found using Breslow estimator \cite{lin2007breslow}, with the relative hazard of that individual [$h_i(t)=h_0(t) \times \psi(X_i)$].
The individual's survival curve is then constructed as $S_i(t)=exp[-H_i(t)]$ where $H_i(t)$ is the cumulative sum of $h_i(t)$ over time $t$. This survivor function will be similar to the one shown in Figure ~\ref{fig:multiple_survplot}a, but instead of characterizing the population, it would be for the individual. 
In the survivor function, a natural estimate of time-to-open would be to use the median survival time, typically referred to as $t(50)$, which corresponds to the time when the survival probability is $0.5$. Given that the observed overall survival rates are quite high in the current scenario, due to the fact that most marketing emails are never opened, the median might not be defined if the survival curve flattens above $S(t) = 0.5$ on the Y-axis, as in Figure ~\ref{fig:multiple_survplot}b. Thus, the concept is generalized to the notion of a $p^{th}$ percentile such that the predicted time-to-open an email by the $i^{th}$ individual, is the largest $t_i$ for which $S_i(t_i(p)) > 1-\frac{p}{100}$ for a given percentile $p$. Therefore, we can have $S(t(10))=0.9$ and $S(t(90))=0.1$.
Figure ~\ref{fig:multiple_survplot} illustrates how a time corresponding to a given survival probability threshold is identified from a survival curve. In the current paper, we evaluate a range of percentiles $p=\{5, 10, 25, 50, 75, 90\}$ as predictors of time-to-open for future emails to that recipient, and all predictions of time are made at the scale of minutes.

\section{Experiments and Results}

In the earlier sections, we have described how the use case of modeling email interaction data maps to concepts in the area of survival analysis. In the current section, we evaluate the performance of a set of models on the two tasks of interest: (a) predicting the likelihood of a marketing message being opened by a recipient, and (b) estimating the time at which this event is likely to occur. 

Table ~\ref{tblModels} contains the models evaluated in the current paper.
The baseline for the classification task is the historical open rate of the corresponding recipient, and the corresponding time-to-open is a constant equal to the censoring window for the prediction of time-to-open.
Logistic Regression is used for the "open vs not" classification task and Linear Regression is used for the time to open of emails. Models $(2) - (3)$ were trained using an Elastic Net ~\cite{elasticNet} regularization penalty to the corresponding likelihoods. Values of the hyper-parameters $\lambda$ (the strength of the regularization) and $\alpha$ (trading off L1 vs L2) were chosen via the use of the \textbf{Validation} dataset. The equivalent parameters of the GBM - number of trees, learning rate and minimum number of observations for a leaf node - were similarly chosen. 

\paragraph{Features:} The set of features available to us are aggregates of actions that a user can take on emails, for example number of messages received, opened, clicked, and open rate in the past. We also included campaign specific information, like whether the last message was opened and/or clicked and consumer specific features like, how long the consumer has signed up for receiving these email messages. In the current paper, we primarily focused on the methodology, instead of identifying an expansive set of features. If a particular feature has good predictive signal, we expect all the models described here to benefit from it.

\begin{table}[!htbp]
\centering
\caption{List of Models}
\label{tblModels}
\resizebox{\columnwidth}{!}{
\begin{tabular}{@{}l@{}}
\toprule
    \multicolumn{1}{l}{$(1)$ \textbf{B}: Baselines} \\
    \multicolumn{1}{l}{$(2)$ \textbf{LR}: Logistic Regression (Classification)/Linear Regression (Time to Open)} \\
    \multicolumn{1}{l}{$(3)$ \textbf{CPH-L}: CoxPH Model with relative hazard, $\psi(\beta, \mathbf{X}) = exp\big(\beta^TX\big)$} \\
    \multicolumn{1}{l}{$(4)$ \textbf{CPH-G}: CoxPH Model with relative hazard, $\psi(\beta, \mathbf{X})$ from a GBM} \\
    \multicolumn{1}{l}{$(5)$ \textbf{MM}: Mixture Model with Proportional Hazards} \\   
\bottomrule
\end{tabular}
}
\end{table}

\subsection{Evaluation Metrics}

For the binary prediction of whether the email will be opened within time $t$, we calculate the area-under-the-ROC-curve (AUC) for the probabilities calculated from the survivor function. For prediction of the time to open an email, we defined the Mean Relative Absolute Deviation ($MRAD$) between the actual value, $t_i$, for the time-to-open and the predicted value $\hat{t_i}$, and is given by 

\begin{equation}
    \label{eqMRAD}
    MRAD = \frac{1}{N} \sum \limits_{i} \frac{|t_i - \hat{t_i}|}{t_i}
\end{equation}

When we calculate the MRAD over all the individuals in the dataset, we refer to the resulting metric as $MRAD(A)$. We also define $MRAD(O)$ as the version of the metric calculated only over the observed individuals, i.e., all the rows for which $\delta_i = 1$. For both the metrics, $\hat{t_i}$ is estimated by the value where $S_i(\hat{t_i}(p))=1-\frac{p}{100}$ for a given value of $p$, as described in the previous section. 

We compared the $5$ models, as shown in Table \ref{tblModels} on the basis of AUC for classification of which recipients will open the messages and MRAD for time to open the messages. Higher the AUC, better the classifier, and a lower MRAD shows lesser deviance between the observed and the predicted. We now describe different sets of experiments evaluating various aspects of the models.

\subsection{Comparison of Models Across Censoring Windows}

\begin{table*}[t]
\centering
\caption{Comparison of the Models under AUC and MRAD across Censoring Windows}
\label{tblSummary}
\resizebox{\linewidth}{!}{
\begin{threeparttable}
\begin{tabular}{l|lllll|lllll|lllll}
\toprule
               & \multicolumn{5}{l|}{\textbf{Censoring Window = 3 hours}} 
               & \multicolumn{5}{l|}{\textbf{Censoring Window = 6 hours}}  
               & \multicolumn{5}{l}{\textbf{Censoring Window = 12 hours}}  \\
\textbf{Model} & \textbf{B} & \textbf{LR*}  & \textbf{CPH-L} & \textbf{CPH-G} & \textbf{MM} 
               & \textbf{B} & \textbf{LR*}  & \textbf{CPH-L} & \textbf{CPH-G} & \textbf{MM} 
               & \textbf{B} & \textbf{LR*}  & \textbf{CPH-L} & \textbf{CPH-G} & \textbf{MM} \\
\midrule
\textbf{AUC}   & 0.863 & 0.931 & 0.931 & \textbf{0.932} & 0.929 
               & 0.870 & 0.939 & 0.939 & \textbf{0.940} & 0.938 
               & 0.878 & 0.948 & 0.948 & \textbf{0.949} & 0.948 \\
\midrule
\textbf{MRAD(A)}  & 1.226 & 1.332 & 1.085 & 0.941 & \textbf{0.483} 
                  & 2.504 & 1.653 & 1.835 & 1.372 & \textbf{0.678} 
                  & 5.079 & 2.332 & 1.707 & 1.572 & \textbf{1.318} \\
\textbf{MRAD(O)}  & 26.641 & \textbf{8.411} & 11.953 & 12.217 & 9.499 
                  & 40.602 & 11.706 & 23.245 & 14.788 & \textbf{9.832}  
                  & 62.740 & 17.501 & 19.831 & 28.978 & \textbf{15.657} \\
\bottomrule
\end{tabular}
\begin{tablenotes}
\small
\item *AUC results for LR correspond to Logistic Regression. MRAD(A) and MRAD(O) results for LR correspond to Linear Regression.
\end{tablenotes}
\end{threeparttable}
}
\end{table*}

We compared the different models across three different censoring windows $C_i \in \{3, 6, 12\}$ hours, using the predicted time-to-open calculated at $\hat{t_i}(5)$. Section ~\ref{expTimePredict} explores the effect of $p$ on both AUC and MRAD for the various censoring windows. Table ~\ref{tblSummary} summarizes the comparisons, and reports the results for the \textit{best} configuration of parameters on the \textbf{Validation} set. The following observations can be made from Table ~\ref{tblSummary}:
\begin{enumerate}[leftmargin=*]
    \item The separate models for each task - Logistic Regression for classification and Linear Regression for predicting time - are very compelling baselines. The benefit of jointly modeling both tasks via survival models becomes more obvious at the larger censoring windows. As discussed earlier, with $C_i=3$ hours,  $\sim57\%$ of the actual open events are considered censored by the survival models. The noise so introduced reduces with larger choices of $C_i$, allowing \textbf{MM} in particular to perform well
    \item Analyzing the importance of the features in our models,  we find that the historical open rate is expectedly the strongest. AUC numbers when using the open rate as a predictor of a user's future open actions have also been included (\textbf{B})
	\item  \textbf{CPH-G} models non-linear relations between the features and hazard function, and 
	as a classifier is stronger than the other models
	\item The assumptions of \textbf{MM} represent the characteristics of our data most accurately. We see that the AUC is similar to that of \textbf{CPH-L}. Note, however, that the predictions of time provided by \textbf{MM} are more accurate than the alternatives at larger $C_i$
	\item The MRAD(A) is much lower than the MRAD(O) as it considers the accuracy of predicting the time-to-open of the censored individuals as well, which the models do efficiently, as evidenced from the high AUC values
	\item When moving through the censoring window options ($C_i \in \{3, 6, 12\}$ hours), the classification task generally gets \textit{easier}, as indicated by the increasing AUC values. This is because the set of observed individuals gets more complete, and separating these from the individuals that are unlikely to open the email is what the model needs to do
\end{enumerate}
An additional observation from these experiments is regarding the validation metric. Across the different hyper-parameter settings, the model that had the highest AUC on the \textbf{Validation} set was chosen, instead of that with lowest MRAD. The choice of the validation metric dictates the future performance of the model - a setting chosen on the basis of higher AUC does not necessarily achieve a lower MRAD on held-out datasets compared to a parameter combination driven by MRAD. This observation indicates that there is an implicit need to pick one of the tasks as being of higher \textit{priority} - here we have chosen the classification task to have a higher priority. Also, for the rest of the experiments in this paper, to compare models for prediction of time to open, we consider MRAD(O), instead of MRAD(A). This is because AUC already provides a comparison for classification and MRAD(O) helps to compare specifically based on those who have opened the messages.

\subsection{Model Performance Across Survival Thresholds}
\label{expTimePredict}

One of the characteristics of the dataset utilized in the current paper is the high percentage of censored observations. This manifests itself in the fact that the survival curves for individuals routinely flatten well before they reach $S_i(t) = 0.5$, an illustration of this at the population level can be seen in Figure ~\ref{fig:multiple_survplot}. Thus, across the different settings and models evaluated here, using a large value of $p$ for $\hat{y_i} = 1 - S_i(\hat{t_i}(p)) $ provides sufficient fidelity for predicting the probability of opening the email.

Table ~\ref{tblTimePredict} describes how the MRAD(O) changes for all the models as we alter the percentile $p$. The column titles use the conventional notation $\hat{t}(p)$, the equivalent survival probability can be worked out readily. $\hat{t}(5)$ corresponds to the largest time when $S_i(\hat{t_i}(5)) > 0.95$. Note that the AUC values remain constant by design, and therefore are not displayed.
In general, as $p$ is reduced, since we are operating in the flatter regions of the survival curves, all $\hat{t}(p)$ become equal to the duration of the censoring window. The point at which the MRAD(O) values appear to reach the maximum is indicative of where the corresponding saturation region is in the survival plot. We would expect the corresponding MRAD(O) values to approach that of the baseline in Table ~\ref{tblSummary}.

In that sense, \textbf{MM} is most closely modeling the data - there is $\sim95\%$ of the population that never opens their messages and the survival curve for \textbf{MM} arrests its drop around $p=5$ percentile, with maximum MRAD(O) observed for all time points after that. The survival curve for \textbf{CPH-L} on the other hand seems to gradually go towards the same maximum MRAD(O), around $\hat{t}(90)$. The survival curve for \textbf{CPH-G} is in between the \textbf{MM} and \textbf{CPH-L}.

\begin{table}[htb]
\centering
\caption{The effect of varying the percentile $p$ of Survivor Function on the prediction quality of the time-to-event}
\label{tblTimePredict}
\resizebox{\linewidth}{!}{
\begin{tabular}{l|l|llllll}
\toprule
\textbf{Censoring}                          &                {\textbf{Model}}        & \multicolumn{6}{c}{\textbf{MRAD(O)}}  \\
\textbf{Window}         &                 &  \textbf{$\hat{t}(5)$}  & \textbf{$\hat{t}(10)$} & \textbf{$\hat{t}(25)$} 
													& \textbf{$\hat{t}(50)$} & \textbf{$\hat{t}(75)$} & \textbf{$\hat{t}(90)$} \\
\midrule                                                      
\multirow{3}{*}{\textbf{3 hours}}  & \textbf{CPH-L}                &  11.952 & 12.608 & 13.929 & 16.744 & 23.462 & 25.716 \\
                                   & \textbf{CPH-G}                &  12.217 & 14.593 & 18.501 & 25.407 & 26.629 & 26.641 \\
                                   & \textbf{MM}                   &   9.499 & 26.641 & 26.641 & 26.641 & 26.641 & 26.641 \\
\midrule
\multirow{3}{*}{\textbf{6 hours}}  & \textbf{CPH-L}                &  23.245 & 17.456 & 18.857 & 19.870 & 27.041 & 34.542 \\
                                   & \textbf{CPH-G}                &  14.788 & 19.588 & 26.233 & 30.970 & 38.102 & 40.602 \\
                                   & \textbf{MM}                   &  9.832  & 12.483 & 40.602 & 40.602 & 40.602 & 40.602 \\
\midrule
\multirow{3}{*}{\textbf{12 hours}} & \textbf{CPH-L}                &  19.831 & 34.632 & 27.229 & 32.510 & 40.356 & 40.750 \\
                                   & \textbf{CPH-G}                &  28.978 & 31.866 & 29.986 & 34.590 & 57.086 & 61.040 \\
                                   & \textbf{MM}                   &  15.657 & 21.705 & 62.545 & 62.740 & 62.740 & 62.740 \\
\bottomrule
\end{tabular}
}
\end{table}

\subsection{Measuring Model Stability}
\label{expStability}

In this section we test the sensitivity of our models, and of the corresponding fitting processes, to variations in the data. We construct an experiment where we take bootstrap samples of the \textbf{Training} set, fit each model in turn to the datasets generated, and calculate the AUC and MRAD(O) metrics for the various models on the \textbf{Validation} set. The hyper-parameter values used were the optimal parameters for defining classification at $\hat{t_i}(5)$, same as for the results reported in Table ~\ref{tblSummary}. 
The mean and standard deviation of the metrics across the samples, for each of the four models, are shown in Table ~\ref{tblBoostrap}.

\begin{table}[htb]
\centering
\caption{Mean and Standard Deviation of AUC \& MRAD(O) respectively on $10$ boostrapped samples}
\label{tblBoostrap}
\begin{threeparttable}
\begin{tabular}{l|l|llll}
\toprule
\textbf{Censoring}   &    \textbf{Model}                     & \multicolumn{2}{c}{\textbf{AUC}} & \multicolumn{2}{c}{\textbf{MRAD(O)}}        \\
\textbf{Window}  &   &   \textbf{Mean}  & \textbf{StdDev} & \textbf{Mean} & \textbf{StdDev} \\
\midrule                                                      
\multirow{4}{*}{\textbf{3 hours}}  & \textbf{LR*}                   &  0.931         &  4e-5           & \textbf{8.215}    &   0.036         \\
                                   & \textbf{CPH-L}                &  0.931         &  2e-5           &  13.579       &  1.913          \\
                                   & \textbf{CPH-G}                &  0.929         &  8e-3           &  13.746       &  0.743          \\
                                   & \textbf{MM}                   &  0.929         &  3e-4           &  9.277       &   1.854           \\
\midrule
\multirow{4}{*}{\textbf{6 hours}}  & \textbf{LR*}                   &  0.939         &  4e-5           & 11.651    &   0.079         \\
                                   & \textbf{CPH-L}                &  0.939         &  1e-5           &  20.301       &   2.486         \\
                                   & \textbf{CPH-G}                &  0.939         &  3e-4           &  19.208       &   0.798         \\
                                   & \textbf{MM}                   &  0.938         &  2e-4           &  \textbf{9.753}        &   1.194           \\
\midrule
\multirow{4}{*}{\textbf{12 hours}} & \textbf{LR*}                   &  0.948         &  4e-5           & 17.514    &   0.096         \\
                                   & \textbf{CPH-L}                &  0.948         &  7e-5           &   38.143      &    3.333        \\
                                   & \textbf{CPH-G}                &  0.949         &  2e-4           &   29.455      &    1.071        \\
                                   & \textbf{MM}                   &  0.948         &  2e-4           &   \textbf{15.444}      &    1.683         \\  
\bottomrule
\end{tabular}
\begin{tablenotes}
\small
\item *AUC results for LR correspond to Logistic Regression and MRAD(O) results for LR correspond to Linear Regression.
\end{tablenotes}
\end{threeparttable}
\end{table}

This sampling procedure allows us to obtain proxy estimates for the confidence intervals for each metric via the standard deviation. This would be required if we were to reliably order the different models. We note that AUC in general has very low variability, and therefore, the ordering over the models on the basis of AUC is reliably stable. MRAD(O) on the other hand, shows more sensitivity to input variations.
This offers another reason as to why AUC might be a more appropriate metric for validation/model-selection. 

The difference between the two metrics can be attributed to the fact that AUC is a ranking metric that is an aggregate over the entire list of consumers, i.e., even if the fitted model parameters across the samples were slightly different, as long as the ordering imposed by $S_i(t)$ over the dataset does not change appreciably, AUC remains the same. In contrast, MRAD is a comparison between the individual predicted ($\hat{t_i}$) and actual ($t_i$) time-to-open, and is expected to be more sensitive. 

When comparing across the models for a given censoring window, we find that \textbf{MM} provides the lowest mean of the MRAD values, with their standard deviations being higher than that of \textbf{CPH-G}.

\subsection{Out-of-Time Model Evaluation}
\label{expOOT}

The final set of experiments verify if the model continues being predictive into the future. To do this, we use the \textbf{Test} dataset that has been held aside up until now. We combine the \textbf{Training} and the \textbf{Validation} sets, and use this concatenated data for building the models. The parameters that are used as input are those that have been chosen via the validation process. This not only includes the hyper-parameters of the models (e.g. the $\lambda$ and $\alpha$ for the Elastic Net based models), but also the choice of percentile $p$ to be used for predicting time (which is the one with least MRAD(O) in Section ~\ref{expTimePredict} for each model). 

\begin{table}[htb]
\centering
\caption{AUC and MRAD(O) for each of the models for the out-of-time dataset}
\label{tblOOT}
\begin{threeparttable}
\begin{tabular}{l|l|ll}
\toprule
\parbox[t]{0.18\columnwidth}{\textbf{Censoring \\Window}}  
            & \multicolumn{1}{c|}{\textbf{Model}} & \multicolumn{1}{c}{\textbf{AUC}} &\multicolumn{1}{c}{\textbf{MRAD(O)}}  \\
\midrule
\multirow{4}{*}{\textbf{3 hours}}  & \textbf{LR*}                  &  0.937         &   7.753         \\
                                   & \textbf{CPH-L}                &  0.937         &  10.009         \\
                                   & \textbf{CPH-G}                &  0.934         &  13.787          \\
                                   & \textbf{MM}                   &  0.935         &   \textbf{7.381}            \\
\midrule
\multirow{4}{*}{\textbf{6 hours}}  & \textbf{LR*}                  &  0.944         & 10.194         \\
                                   & \textbf{CPH-L}                &  0.944         & 23.227         \\
                                   & \textbf{CPH-G}                &  0.940         &   18.339         \\
                                   & \textbf{MM}                   &  0.942         &   \textbf{9.340}      \\
\midrule
\multirow{4}{*}{\textbf{12 hours}} & \textbf{LR*}                  &  0.952         &   13.409         \\
                                   & \textbf{CPH-L}                &  0.952         &   29.131        \\
                                   & \textbf{CPH-G}                &  0.950         &    21.080        \\
                                   & \textbf{MM}                   &  0.951         &    \textbf{11.653}         \\  
\bottomrule
\end{tabular}
\begin{tablenotes}
\small
\item *AUC results for LR correspond to Logistic Regression and MRAD(O) results for LR correspond to Linear Regression.
\end{tablenotes}
\end{threeparttable}
\end{table}

The results presented in Table ~\ref{tblOOT} indicate that the performance does not degrade over time. In all the models, the AUC values are high.
As was seen in Section ~\ref{expTimePredict}, with the right choice of the percentile $p$ to be used for estimating the time-to-open, the Mixture Model obtains prediction accuracy that is better than the other models. Coupled with the fact that the model assumptions of \textbf{MM} aligns more closely with the characteristics of our data, we believe that designing better optimization strategies with this model would be a fruitful avenue to explore in future work.

\section{Discussion and Conclusion}
\label{secDiscussion}
In this paper, we proposed the use of survival models for the problems of predicting if a recipient will open an email or not, and estimating the corresponding time-to-open. We have provided a step by step procedure to check model assumptions using statistical tests. These tests help to determine whether the models that we have compared are meaningful for the data that we have. We have then performed a thorough experimental comparison of the models. Since a large proportion of emails are not opened, the median time-to-open is not defined and hence cannot be used as the estimate of future time-to-open. We therefore tested our models for various percentiles and the $5^{th}$ percentile is found to be the best predictor for time-to-open.

Across our experiments, it was observed that for the task of predicting if an individual will open the email or not, a traditional classification based model (here, Logistic Regression) is a strong baseline. Survival analysis based models achieve similar accuracy to the classifier when predicting the event. While the linear regression is a strong baseline for predicting time when the censoring window is $3$ hours, the proposed survival analysis performs best with the addition of new data when the censoring window is larger.
Focusing on marketing email messages, we observe that a large fraction ($\sim90\%$) of emails do not get opened (even after $10$ days), owing to the fact that certain recipients are inclined to ignore such messaging. This characteristic is not commonly encountered in applications of survival analysis. Given data of this nature, a mixture model that allows for individuals to have very low predispositions to experiencing the event (here, opening the email) was explored. This differential modeling of sub-populations was done in association with a Proportional Hazards assumption and was shown to be most effective. As future extensions to our work, we plan to derive a regularization framework for the mixture model, which promises to be more effective specially when the model includes a large number of features with correlations between them.

\begin{acks}
We thank Frederic Mary in Adobe Product Management for his insights, and anonymous reviewers for their constructive feedback.
\end{acks}

\bibliographystyle{ACM-Reference-Format}
\bibliography{sigproc} 

\end{document}